\documentclass[runningheads]{llncs}
\usepackage{graphicx}

\usepackage{tikz}
\usepackage{comment}
\usepackage{amsmath,amssymb} 
\usepackage{color}
\usepackage[T1]{fontenc}
\usepackage[accsupp]{axessibility}  

\usepackage[width=122mm,left=12mm,paperwidth=146mm,height=193mm,top=12mm,paperheight=217mm]{geometry}

\DeclareMathOperator{\diag}{diag}

\newcommand{\bs}{\mathbf{s}}
\newcommand{\bz}{\mathbf{z}}

\newcommand{\bx}{\mathbf{x}}
\newcommand{\bX}{\mathbf{X}}
\newcommand{\bh}{\mathbf{h}}

\newcommand{\bM}{\mathbf{M}}

\usepackage{float}
\usepackage{caption}
\usepackage{subcaption}

\begin{document}
\pagestyle{headings}
\mainmatter
\def\ECCVSubNumber{11}  
\title{Rethinking Trajectory Prediction via ``Team Game''}

%
\author{Zikai Wei\inst{1} \orcidID{0000-0002-8749-2188} 
\and
Xinge Zhu \inst{1} \orcidID{0000-0003-0107-8099} 
\and 
\\ Bo Dai \inst{2} \orcidID{0000-0003-0777-9232} 
\and
Dahua Lin \inst{1} \orcidID{0000-0002-8865-7896} 
\
}
\authorrunning{Z. Wei et al.}
%
%
\institute{The Chinese University of Hong Kong\\
\email{\{wz018,xg018, dhlin\}@ie.cuhk.edu.hk} 
 \and
Shanghai AI Laboratory \\
\email{daibo@pjlab.org.cn}}

\maketitle

\begin{abstract}
To accurately predict trajectories in multi-agent settings, e.g. team games, it is important to effectively model the interactions among agents. Whereas a number of methods have been developed for this purpose, existing methods implicitly model these interactions as part of the deep net architecture. However, in the real world, interactions often exist at multiple levels, e.g. individuals may form groups, where interactions among groups and those among the individuals in the same group often follow significantly different patterns. 
In this paper, we present a novel formulation for multi-agent trajectory prediction, which explicitly introduces the concept of interactive group consensus via an interactive hierarchical latent space. This formulation allows group-level and individual-level interactions to be captured jointly, thus substantially improving the capability of modeling complex dynamics. 
On two multi-agent settings, \emph{i.e.} team sports and pedestrians, the proposed framework consistently achieves superior performance compared to existing methods. 
\end{abstract}

\begin{figure}[h]
  \vspace{-0.4cm}
  \hsize=\textwidth
  \centering
   \begin{subfigure}[b]{0.32\textwidth}
         \includegraphics[width=\textwidth]{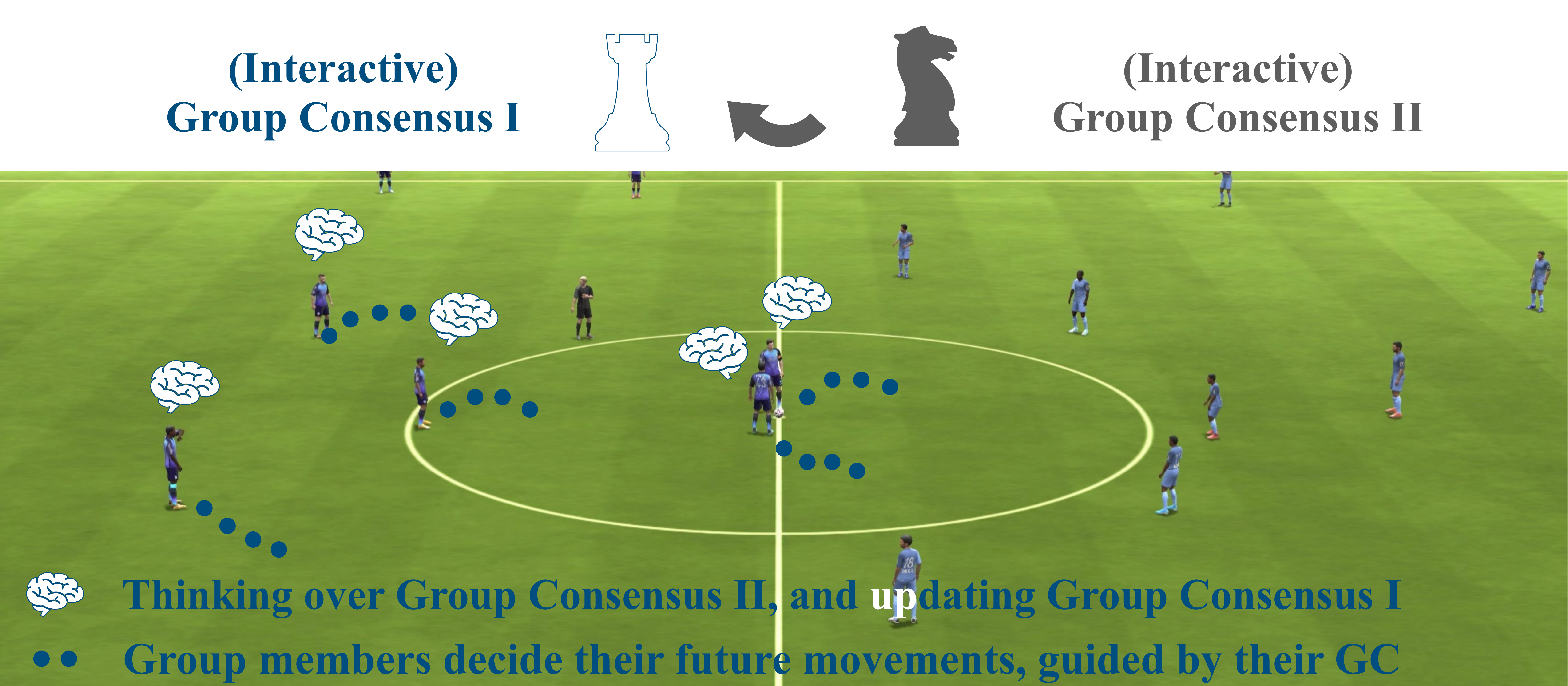}
         \caption{Conjecture on the group consensus of Group II.} 
         \label{fig:idea_a}
    \end{subfigure}
    \begin{subfigure}[b]{0.32\textwidth}
         \includegraphics[width=\textwidth]{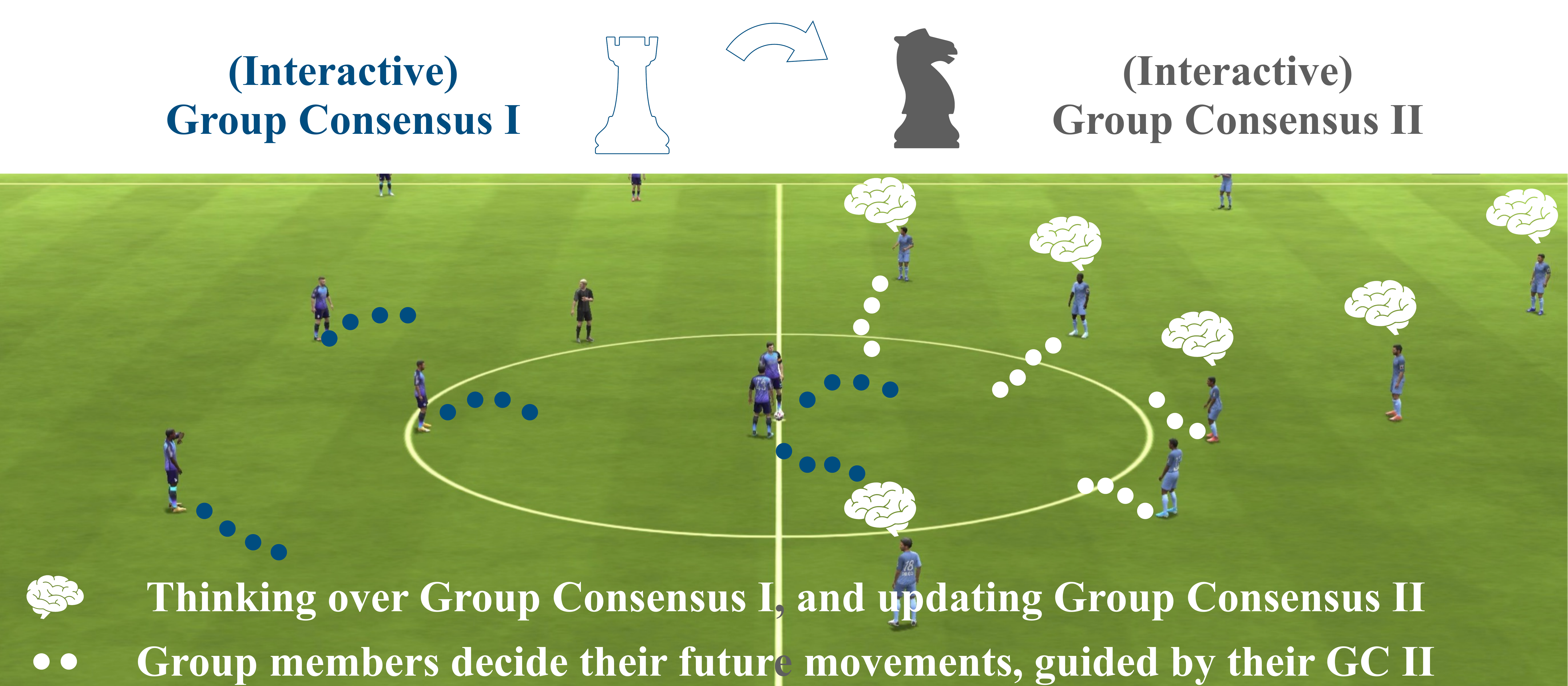}
         \caption{Reasoning on the group consensus of Group I.}
         \label{fig:idea_b}
     \end{subfigure}
    \begin{subfigure}[b]{0.32\textwidth}
         \includegraphics[width=\textwidth]{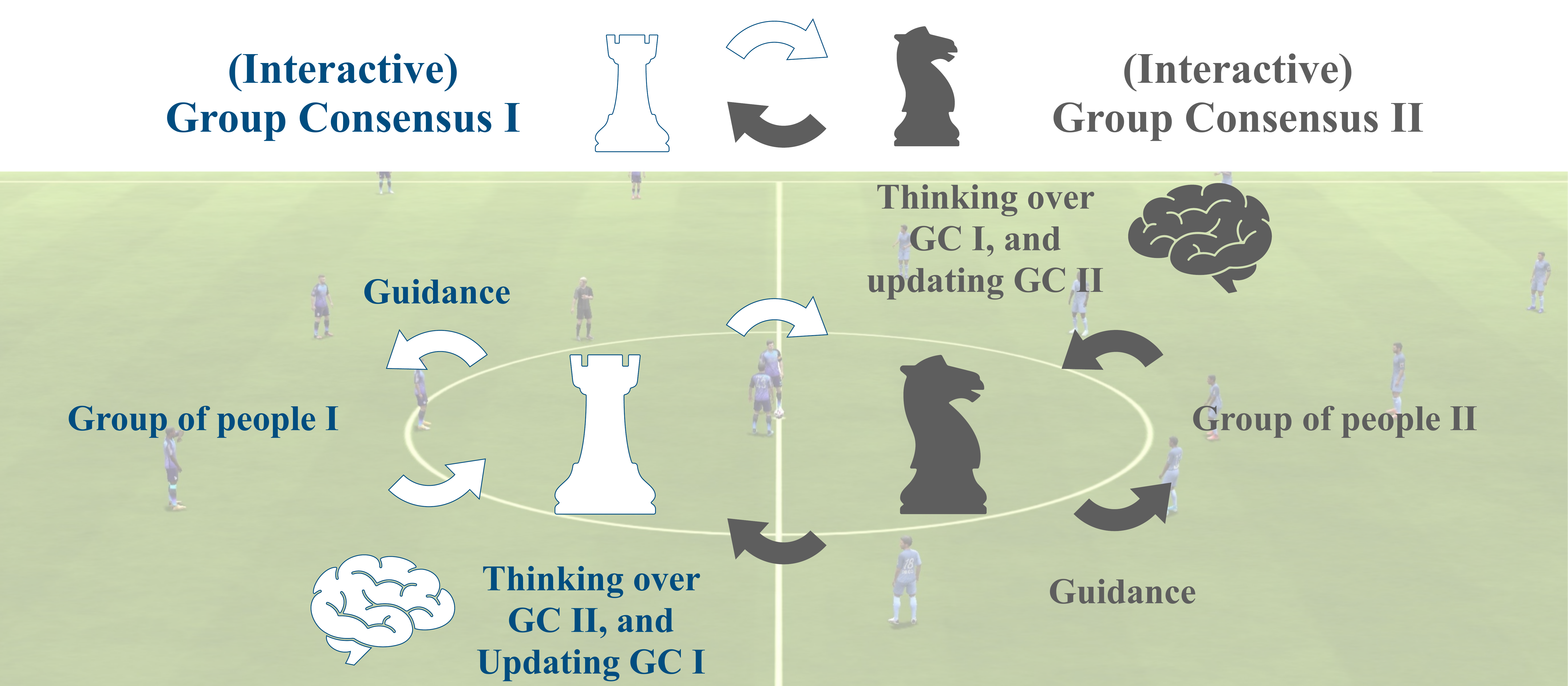}
         \caption{The process of Interactive Group Consensus}
         \label{fig:idea_c}
    \end{subfigure}
  \vskip -0.2cm
  \caption{\textbf{The concept of interactive group consensus}. (a) The members from Group I formulate their group consensus/strategy (GC) by  conjecturing about their opponent's group strategy. Later, their GC I guides individuals' preferences and further influences their future trajectories. (b) The GC II is formulated in the similar way and it later unifies the goals of its group members. (c) The Interactive Group Consensus (IGC) summarizes the inter-dependence between different group consensuses (strategies). 
}
  \label{fig:idea}
\end{figure}


\section{Introduction} \label{introduction}

Predicting the future trajectory of a person in a certain environment is one of the fundamental topics for human behavior analysis, which attracts a lot of attention across computer vision, sociology, and cognitive science. 
The development of human trajectory prediction also benefits a lot of downstream tasks including anomaly detection, crowd analysis, as well as team sports analysis. 
In recent years, learning-based methods~\cite{zhang2019sr,huang2019stgat,fang2020tpnet,sun2020recursive,hauri2020multi} 
have made remarkable signs of progress in trajectory prediction, which significantly push forward the state of the art to a new level. 
However, how to explicitly model interactions at multiple hierarchies is complicated and remains open. Another neglected question is how to involve the collaboration and competitiveness among multiple agents in human trajectory prediction.

To tackle these issues, we reformulate the trajectory prediction problem from the perspective of game theory (sequential game) and sociology: agents who have the same interests may 1) form a group, 2) establish a group consensus (strategy) by conjecturing on other groups' strategies, and 3) take actions guided by their group consensus. We introduce the \emph{interactive group consensus} as a new concept to summarize this ``team game" process.

\begin{figure}[t]
  \centering
     \begin{subfigure}[b]{0.23\textwidth} 
         \centering
         \includegraphics[width=\textwidth]{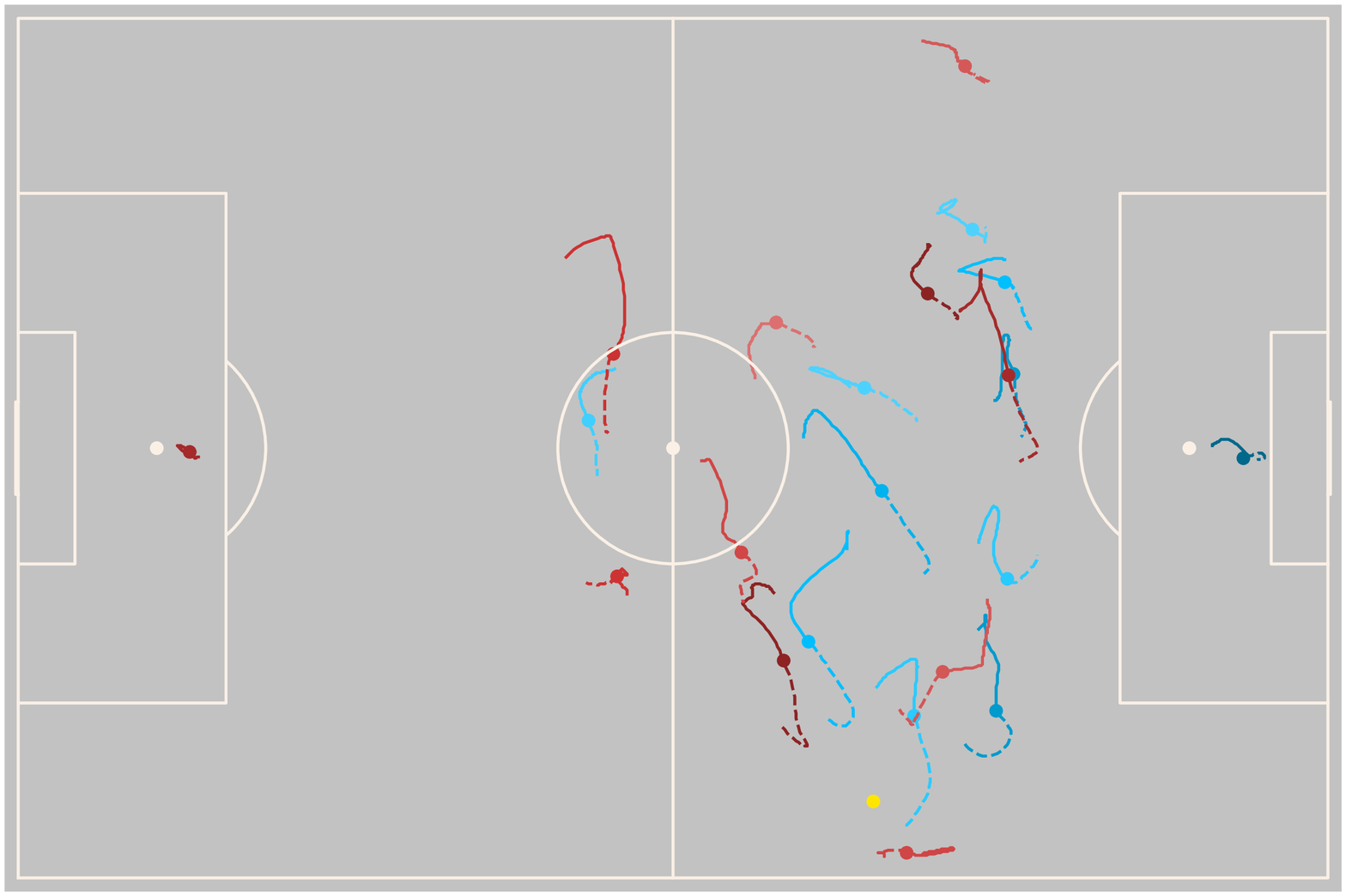}
        \vskip -0.2cm
         \caption{GT Trajectories}
     \end{subfigure}
     \hfill
     \begin{subfigure}[b]{0.23\textwidth}
         \centering
         \includegraphics[width=\textwidth]{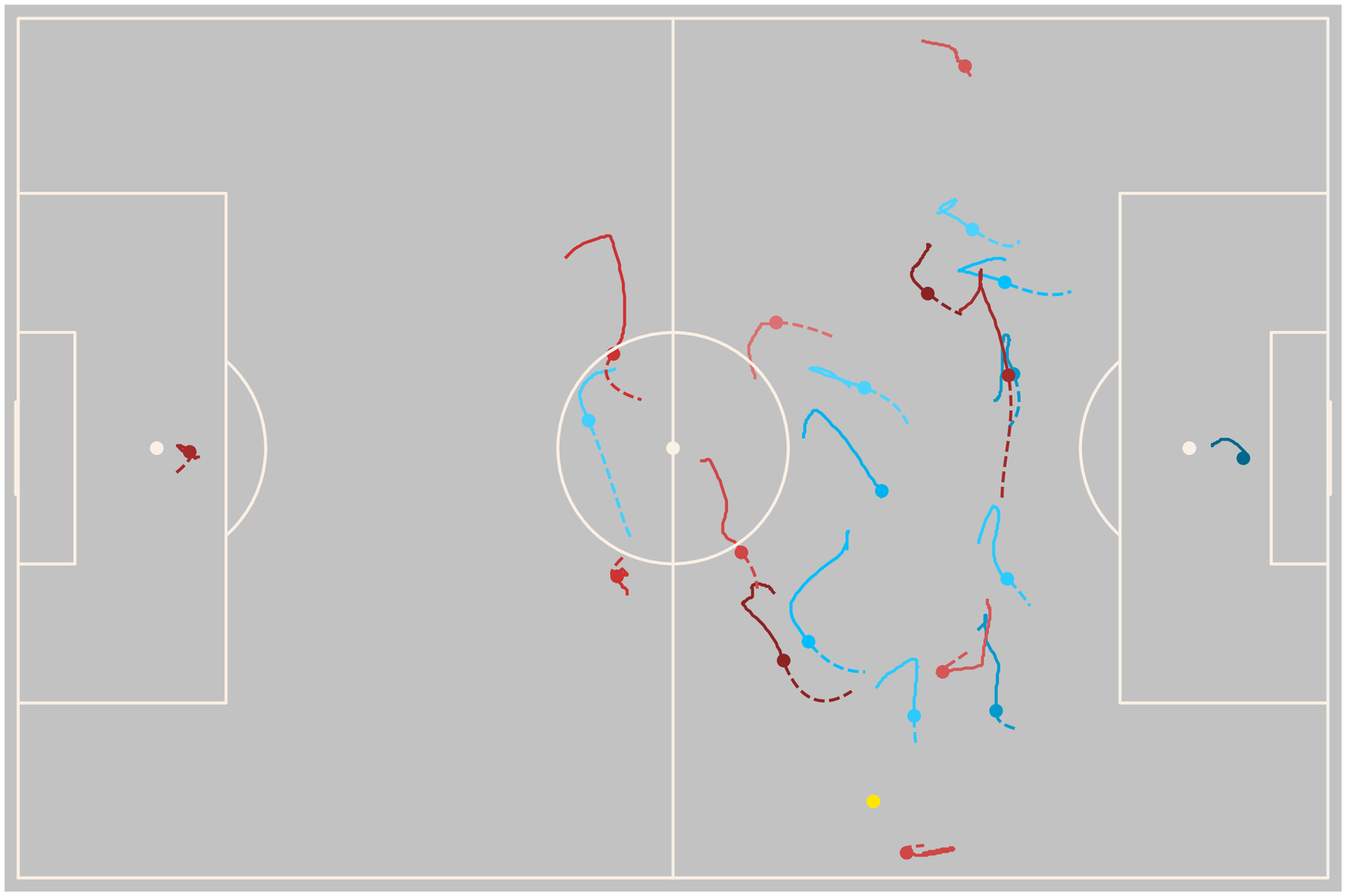}
        \vskip -0.2cm
        \caption{Ours with IGC}
     \end{subfigure}
     \hfill
     \begin{subfigure}[b]{0.23\textwidth}
         \centering
         \includegraphics[width=\textwidth]{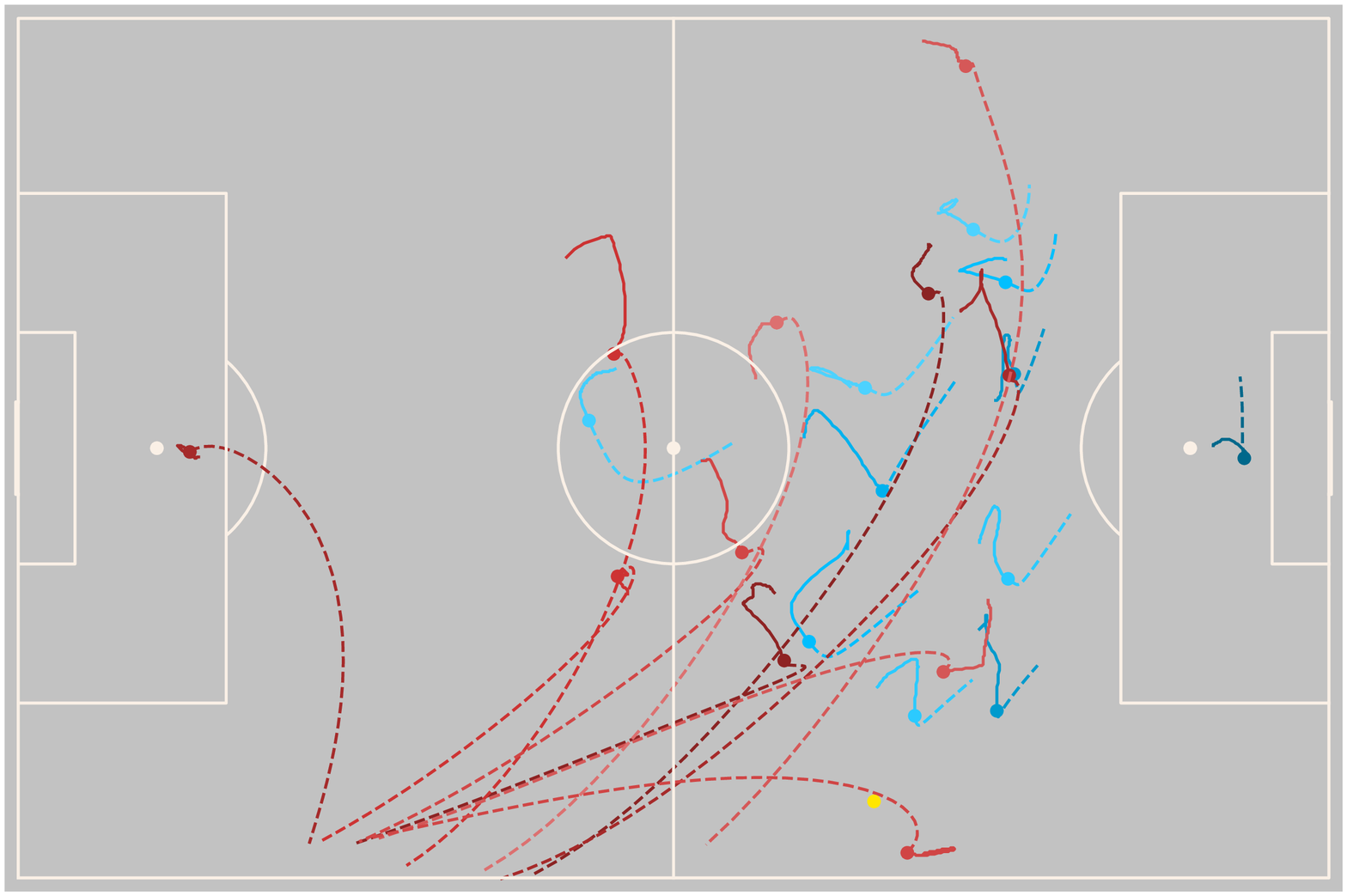}
        \vskip -0.2cm
         \caption{No DEF}
     \end{subfigure}
     \hfill
     \begin{subfigure}[b]{0.23\textwidth}
         \centering
         \includegraphics[width=\textwidth]{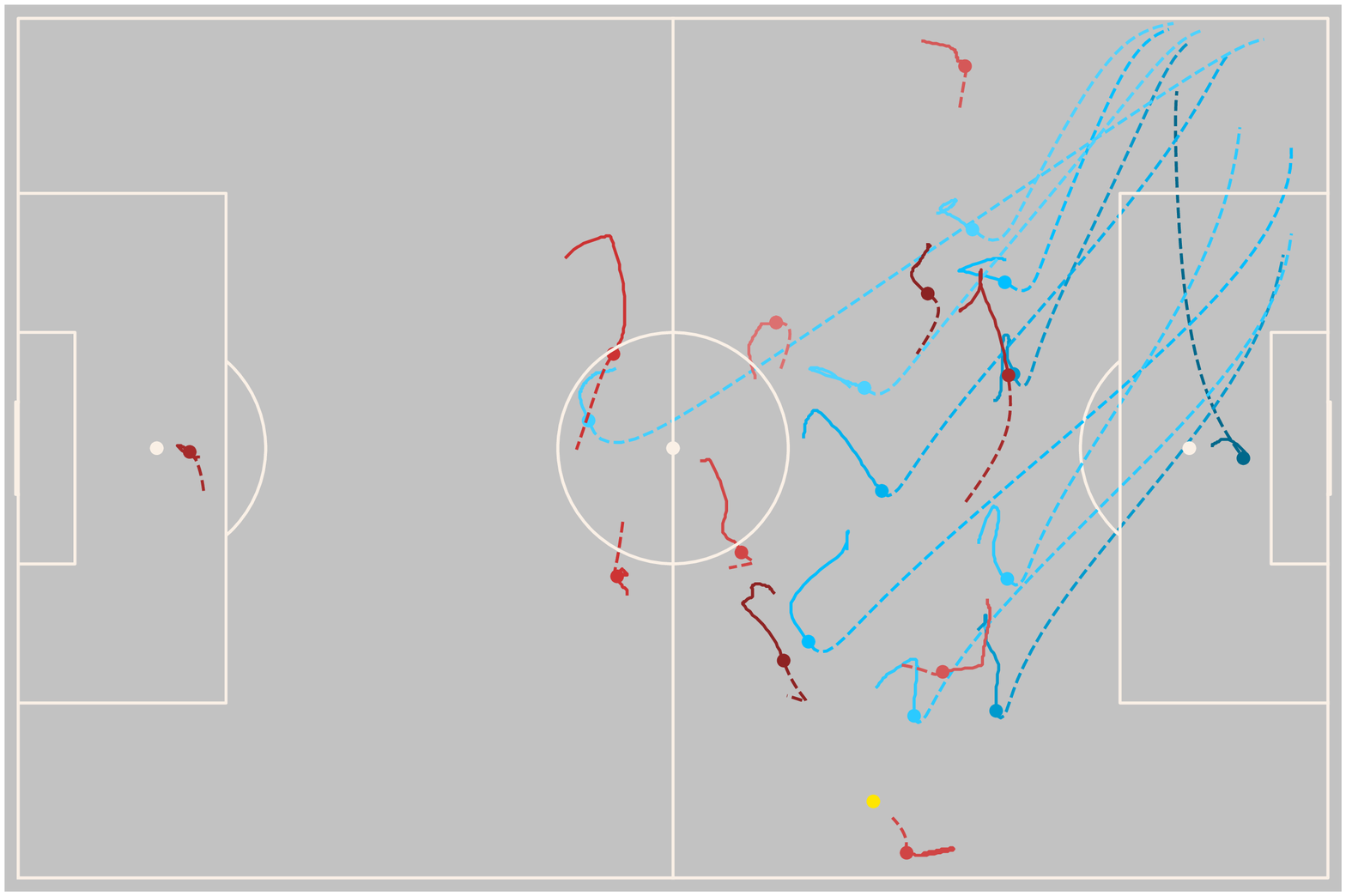}
         \vskip -0.2cm
         \caption{No OFF}
     \end{subfigure}
  \caption{\textbf{The importance of conjecturing about interactive group consensus (IGC).}  The trajectories in \textcolor{red}{red} and \textcolor{blue}{blue} represent the \textcolor{red}{offensive} and \textcolor{blue}{defensive} players respectively.
  (a) represents the ground-truth trajectories. (b) represents the trajectories generated with IGC. (c) shows the trajectory prediction without guessing about group consensus of the defensive team (DEF). (d) shows a case without conjecture about the offensive team (OFF).  
}
  \label{fig:importance-IGC}
  \vspace{-0.2cm}
\end{figure}

The importance of interactive group consensus can be observed from the fact that people naturally establish different groups and formulate their \emph{group consensus} across the whole group, as shown in Figure \ref{fig:idea}: 
(i) Internally, the group consensus (formulated by all group members via negotiation) unifies the preferences or behaviors of people in the same team. (ii) Externally, strategies or group consensuses among different groups have a mutual impact on each other. 
In cases like \emph{team sports}, 
during the formation of team strategy, 
all players in a team contemplate 
how to gain and hold a competitive advantage over their opponents. Once their team strategy is explicitly created, 
the whole team will unify their behaviors to
achieve their common goal. 
Consequently, a player’s movement is not only related to his individual motivation but also his team strategy and the strategy of the opposing team.
In other cases like \emph{pedestrians}, the process of ``Team Game" also exits: neighbors moving in the same direction may implicitly form a group and follow approximately the same logic to prevent collisions with other group of people.
Neglecting interactive group consensus, some failure cases occur, see Figure \ref{fig:importance-IGC}. 
Given the importance of incorporating the interactive group consensus in human trajectory prediction, 
in this paper, we introduce the Spatial-Temporal Graph \emph{V}ariational \emph{RNN} with an \emph{I}nteractive \emph{H}ierarchical Latent Space, \emph{IHVRNN} for short,
a novel framework for human trajectory prediction,
that explicitly takes interactive group consensus into consideration.


Our contribution can be summarized as follows: 
To the best of our knowledge, this is the first work to introduce the interactive group consensus as well as how it helps modeling multi-level dynamics in human trajectory prediction.
To fully investigate the effectiveness of interactive group consensus, we propose a novel variational recurrent neural network that can jointly estimate multiple levels of interactions for multi-agent scenarios. Our method consistently achieves improved performance on both team sports (Europa League dataset \cite{bialkowski2014win,le2017coordinated}) and pedestrian prediction (ETH \cite{park2018sequence} and UCY \cite{lerner2007crowds} datasets).

\section{Related work}
\label{related_work}

Team sports and pedestrians are two widely explored multi-agent scenarios for trajectory prediction. 

\vspace{1ex}
\noindent{\textbf{Pedestrian.}}~~~
Recent works on pedestrian trajectory prediction mainly focus on how to model the interaction among pedestrians and aggregate the spatial impact across a large number of people~\cite{liang2019peeking,li2019way,sadeghian2019sophie,huang2019stgat,mohamed2020social,salzmann2020trajectron++}. 
The interaction among pedestrians is first studied by social pooling methods~\cite{alahi2016social,gupta2018social,R1_bisagno2018group}. Later, introduces a better way to summarize the spatial interaction between multiple agents via graph attention networks \cite{cho2015describing,velivckovic2017graph}. 
To capture the relative importance of each pedestrian among a crowd, \cite{vemula2018social} formulates the problem of human trajectory prediction as a spatial-temporal graph. \cite{R1_bisagno2018group} models individual interactions within a group. \cite{mangalam2020not,mangalam2021goals} use goal-based methods that propose possible endpoints for each pedestrians and then generate multiple trajectories based on these endpoints.
However, these methods only consider agents' interaction at an individual level, neglecting their interactions at multiple levels. 

\vspace{1ex}
\noindent{\textbf{Team~Sports.}}~~~
Imitation learning \cite{le2017coordinated,liu2019emergent,kurach2019google} learn policies and demonstrations from multi-agent trajectories. 
{Variational Autoencoder} (VAE) \cite{kingma2013auto} and Conditional VAE \cite{sohn2015learning} provide a springboard for  sequential generative models. \cite{zhan2018generating} uses sequential generative models in a hierarchical structure to learn multi-agent trajectories of basketball gameplay. 
\cite{sun2019stochastic} proposes a graph-structured variational recurrent neural network for current state estimation and future state prediction based on ambiguous visual information. 
Besides, \cite{hauri2020multi} proposes a multi-model trajectory prediction of NBA players in one team and \cite{hauri2020multi} consider two teams. 
However, they fail to take the interactive group consensus into consideration, which is crucial in team sports prediction\footnote{We note that, as far as we know, our paper is the first to model the interactive group consensus.}.


{\color{black}  
\vspace{1ex}
\noindent{\textbf{Crowd~Analysis.}}~~~Crowd statistics and behavior understanding are two main research areas in crowd analysis \cite{R2_grant2017crowd}. The works in behavior understanding as related to crowds and these works mainly focus on identifying groups \cite{R7_ge2012vision,R10_zhong2015learning,R11_zhou2012coherent} and interactions of individuals \textbf{within} small groups \cite{R3_kok2016crowd,R4_li2014crowded,R5_allain2009crowd,R6_zhou2015learning}. 
Our idea differs from these works in the following two aspects: 
First, in our work, a group is defined as a set of individuals of interest whose members may have distinct behaviors, which is not the same as clustering behaviors  \cite{R1_bisagno2018group,R8_ryoo2011stochastic,R9_yi2015understanding}. 
Second, although these works divide individuals into different groups and model interactions of individuals within groups, they still tackle the interactions modeling problem at the individual level, \emph{i.e.}, they fail to capture interdependence between a group and its group members and ignore the mutual influences among different groups. 
} 


\section{Methodology}

To fully explore the process of team game, we introduce a novel variational framework with a two-level hierarchical latent space, where this latent space encodes the collaboration and competition among multiple agents.

\subsection{Problem formulation}
Let $\bx_t^{(i)} \in \mathcal{R}^2$ denote the 2-dimensional location of agent $i$ at time $t$, and let  $( \bx_{t-l+1}^{(i)}, \bx_{t-l+2}^{(i)}, \dots,\bx_{t}^{(i)})$ be a corresponding trajectory with time horizon length $l$. Let $\bX = \{ \bx^{(1)},  \bx^{(2)}, \dots, \bx^{(n)}\}$ be an unordered set of trajectories, corresponding to one segment of a scene, where $n$ is the number of agents in this segment. In practice, multiple agents, such as pedestrians and team players, can be divided into different groups at the beginning. 
Generally, $\bX$ can be divided to $N_g$ groups without intersection, \emph{i.e.}, $\bX=\{\bx^{1}, \dots, \bx^{N_g}\}$. Let $\mathcal{G}$ be the index set of these $N_g$ groups.
It is trivial to divide all players into two static groups for team sports. While for pedestrian prediction, the pedestrians' membership changes \textit{dynamically} as their walking directions or neighborhoods change. For the consistency of interactive group consensus, we assume there are $N_g$ groups in a scene, one of which can be an active or inactive group, where an inactive group means no team member inside it. The number of active groups changes as pedestrians' membership changes. Therefore, we only update group consensus for the active group(s). 
For simplicity, we denote $\bx_t = \{\bx_{t}^{(i)}, i =1, ..., k\}$ as the current locations of all agents from the current group of interest.

\vspace{1ex}
\noindent{\textbf{Interactive Group Consensus.}}~
Let $\bs_t$ denote the group consensus of a group of interest, and $\bs_t^o, o \in \mathcal{G}$ denote the other group(s)' group consensuses\footnote{More exactly, we assume that group $i$ is current group of interest among $N_g$ groups and its group consensus is denoted as $\bs_t^i$, while the other group(s)' consensuses/strategies can be denoted as $\bs_t^o=\{\bs_t^j, j=1,\dots,N_g; j\neq i\}$. To simplicity of the notation, we use $\bs_t$ and $\bs_t^o$ instead.}, where $\mathcal{G}$ is the index set of $N_g$ groups without intersection.
The interactive group consensus of current group of interest can be defined as the inter-dependence between $\bs_t$ and $\bs_{t-1}^o$, which can be captured by
$p(\bs_t,\bs_{t-1}^o) = p(\bs_t|\bs_{t-1}^o) p(\bs_{t-1}^o)$. In the context of implementation, we define 
the dynamics of interactive group consensus as $\bs_t|\bs_{<t},\bs_{<t}^o$, and define a group consensus as $\bs_t \sim p(\bs_t|\bs_{<t},\bs_{<t}^o)$.

\begin{figure*}[h]
    \centering
    \includegraphics[width=1\linewidth]{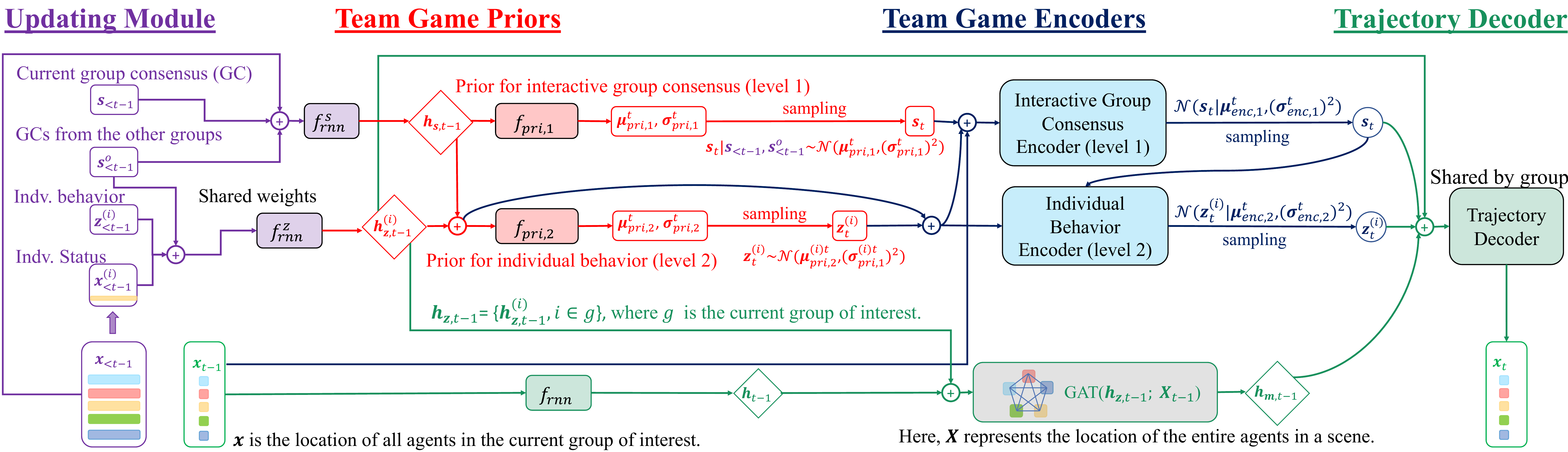} 
    \caption{The pipeline of our IHVRNN from a group perspective. 
    The \textit{updating module} summarizes the past information of 
    interactive group consensus and individual behaviors.
    Later, the \textit{team game priors} show how a group consensus (strategy) is established by conjecturing on other groups' strategy and how it affects its group members' behaviors or actions.
    Further, the \textit{team game encoders} explicitly formulates the dependencies between group members' trajectories, individual behaviors, and different groups' strategies.   
    Additionally, a GAT \cite{velivckovic2017graph} summarizes the social impact from the other people for everyone. Lastly, the \textit{trajectory decoder} decodes all people's trajectories for every person in a scene, based on their group consensus, personal preferences and personal impacts from the other people.
    }
    \label{pipeline}
\end{figure*}

\subsection{Modeling Interactive Group Consensus}
\label{IHVRNN}
After introducing the various interactions and their implementation in our method, we give the whole pipeline of our model (as shown in Figure \ref{pipeline}), where four major components are included. Specifically, the \textit{updating module} summarizes the past common strategies and individual preferences at each time step in their corresponding hidden states. Later, \textit{team game priors} characterize the interdependence among the group consensuses from different groups and specify the dependence of individual preference on the common strategy of her group. Further, the \textit{encoders} map the observed trajectories of the group of interest into an interactive hierarchical latent space (characterizing interactive group consensus, individual preference, and their mutual dependency). Lastly, the sharing \textit{trajectory decoder} will generate or predict the realistic trajectories of all people in a scene. 
In addition, a \textit{refinement module} is introduced to refine the modeling of interaction among individuals, regarding their scope-changing social attention.

\vspace{1ex}
\noindent{\textbf{Updating~Module.}}~~~
The updating module summaries the past information for individual preferences and their group consensus via the following updating functions:
\begin{align}
\bh_{\bz,t}^{{(i)}} & = f_{rnn}^{\bz} ( \bx_{t}^{(i)}, \bz_{t}^{(i)},\bs_{t}, \bh_{\bz,t-1}^{{(i)}}, \bh_{\bs,t-1}),\\
\bh_{\bs,t} &= f_{rnn}^{\bs} ( \bx_{t}, \bs_{t}, \bs_{t}^o, \bh_{\bs,t-1}).
\end{align}
Here, $f_{rnn}^{\bz}$ is a recurrent neural network (RNN) with shared weights. $f_{rnn}^{\bs}$ is a RNN, where its hidden state $\bh_{\bs,t}$ summarizes the past interactive group consensus among the current group and the other group(s) ($\bs_{<t}$ and $\bs_{<t}^o$) from the current group's perspective. Meanwhile, $\bh_{\bz,t}^{{(i)}}$ contains not only a person's past personal information (track $\bx_{<t}$ and individual preference $\bz_{<t}$) but also her group consensus $\bs_{<t}$. 

\vspace{1ex}
\noindent{\textbf{Team~Game~Priors.}}~~~
The \emph{intuition} about the priors comes from: human's preferences (behaviors) and decision making ($\bz_t$) are highly driven by their group consensus (team strategy) $\bs_t \sim p(\bs_t|\bs_{<t},\bs_{<t}^{o})$, where $\bs_{t}^{o} \sim p(\bs_{t}^{o}|\bs_{<t}^{o},\bs_{<t})$. Thus, we design a two-level hierarchical latent space for latent variables $\bs_t$ and $\bz_t$, where group consensuses among different groups are mutually dependent on each other (in particular, \emph{i.e.} $\bs^{i}|\bs^{j}$, $ j = 1,\dots, N_g$ with $i \ne j$). Thus, the conditional priors are defined as 
\begin{align}
    p_{\theta_1}\left(\bs_{t}|\bx_{<t}, \bs_{<t}, \bs_{<t}^o\right) & = \mathcal{N}\left(\bs_{t}|\pmb{\mu}^{t}_{p1 },(\pmb{\sigma}_{p1}^{t})^2 \right),\\
    p_{\theta_2}\left(\bz_{t}^{(i)}|\bx_{<t}^{(i)}, \bz_{<t}^{(i)}, \bs_{<t}\right) & = \mathcal{N}\left(\bz_{t}^{(i)}|\pmb{\mu}_{p2}^{(i),t},(\pmb{\sigma}_{p2}^{(i),t})^2 \right)\\
    \pmb{\mu}_{p1}^{ t},\pmb{\sigma}_{p1}^{t} = f_{p1} & \left(\bh_{\bs,t-1}; \theta_1 \right),\\
    \pmb{\mu}_{p2}^{(i), t},\pmb{\sigma}_{p2}^{(i), t} = f_{p2} & \left(\bh_{\bz,t-1}^{(i)}, \bh_{\bs,t-1}, \bs_{t}; \theta_2\right),
\end{align}
where $f_{p1}$ and $f_{p2}$ are deep neural networks formulating the conditional priors of a two-level hierarchical latent space with learnable parameters $\theta_1$ and $\theta_2$. $\mathcal{N}(\cdot, \pmb{\mu}, \pmb{\sigma}^2)$ denotes a multivariate normal distribution with mean $\pmb{\mu}$ and covariance matrix $\diag(\pmb{\sigma}^2)$. It is worthy to note that covariance matrix with only diagonal elements (instead of a full matrix) can save computational power, which has been proved to be effective in \cite{kingma2013auto}. More theoretical analysis about ``Conditional Priors" is given in supplementary materials.

\vspace{1ex}
\noindent{\textbf{Encoders~with~Interactive~Hierarchical~Latent~Space.}}
The team game encoders map the observed trajectories of a particular group of people $\bx_{\le t}^{(i)}$ to an interactive hierarchical latent space of group consensus (common strategy) $\bs_{t}$ and individual behaviors $\bz_{t}$: 
\begin{align}
    &q_{\phi_1}\left(\bs_{t}|\bx_{<t}, \bz_{<t}, \bs_{<t}, \bs_{<t}^o\right)  = \mathcal{N}\left(\bs_{t}|\pmb{\mu}_{e1}^{ t},(\pmb{\sigma}_{e1}^{t})^2 \right), \\
    & q_{\phi_2}\left(\bz_{t}^{(i)}|\bx_{\le t}^{(i)}, \bz_{<t}^{(i)}, \bs_{<t}\right)  = \mathcal{N}\left(\bz_{t}^{(i)}|\pmb{\mu}_{e2}^{(i), t},(\pmb{\sigma}_{e2}^{(i), t})^2 \right)\\
    &\pmb{\mu}_{e1}^{ t},\pmb{\sigma}_{e1}^{t} = f_{e1}  \left(\bx_{t},\bz_{t-1}, \bh_{\bz,t-1}, \bh_{\bs,t-1}; \phi_1\right),\\
    &\pmb{\mu}_{e2}^{(i), t},\pmb{\sigma}_{e2}^{(i), t} =f_{e2}  \left(\bx_{t}^{(i)}, \bh_{\bz,t-1}^{(i)} ; \phi_2\right)
\end{align}
where $f_{e1}$ and $f_{e2}$ are the corresponding encoder with learnable parameters $\phi_1$ and $\phi_2$ for group consensus and individual behavior respectively. The interactive group consensus encoder $f_{e1}$ concatenates the latest individual status $\bx_{t}$, their individual behaviors $\bz_{t-1}$ and the past summarized information ($\bh_{\bz,t-1}, \bh_{\bs,t-1}$), and then maps them to the \emph{level-1} latent space for group consensus. The individual behavior encoder $f_{e2}$ further maps its individual status $\bx_{t}^{(i)}$ and the hidden state $ \bh_{\bz,t-1}^{(i)}$ to the \emph{level-2} latent space for individual behavior, where the hidden state $ \bh_{\bz,t-1}^{(i)}$ summarizes 
its past individual behavior as well as its past group consensus.

\vspace{1ex}
\noindent{\textbf{Trajectory~Decoder.}}~~~
To incorporate the social interaction from other agents, $f_{d}$ is equipped with $\bh_{\mathbf{m}, t-1}^{(i)}$ which summarizes the other agents' personal impact on agent $i$ via graph attention networks (GAT) \cite{velivckovic2017graph}. Then, the trajectory decoder is defined as
\begin{align}
&p_{\theta}\left(\bx_{t}^{(i)}|\bx_{< t}^{(i)}, \bz_{\le t}^{(i)}, \bs_{\le t}, \bs_{<t}^o \right)  =  \mathcal{N}\left(\bx_{t}^{(i)}|\pmb{\mu}_{d,t}^{(i)},(\pmb{\sigma}_{d,t}^{(i)})^2 \right)\\
&\pmb{\mu}_{d,t}^{(i)},\pmb{\sigma}_{d,t}^{(i)}=f_{d}\left( \bz_{t}^{(i)}, \bs_{t}, \bh_{\bz,t-1}^{(i)}, \bh_{\bs,t-1}, \bh_{\mathbf{m}, t-1}^{(i)}; \theta \right) 
\end{align}
where $\bh_{\bz,t-1}^{(i)}$ and $\bh_{\bs,t-1}$ summarize the past information of individual behavior and inter-group dynamics, respectively. The decoder $f_{d}$ is shared by all agents in a scene with learnable parameters $\theta$.


\vspace{1ex}
\noindent{\textbf{Refinement~Module.}}~~~For modeling the individual-individual interaction via GAT, an agent will pay different attention to its surroundings regarding their distances. To this end, we propose $K$ scope-changing social masks, $\bM_{n \times n}^{[k]},k=1,\dots,K$, to allocate dynamic weights, where $\bM_{n \times n}^{[k]}$ is a symmetric matrix and defined as 
\begin{align}
\bM_{ij}^{[k]} = \bM_{ji}^{[k]}
\begin{cases}
    1, & \text{if}\ d_{ij}\le k \cdot d_{m} \\
    0, & \text{otherwise}
\end{cases},\text{~for~}k < K.
\end{align}
Here, $d_m$ is a distance threshold of attention gap, and $d_{ij}$ is the distance between the agent $i$ and agent $j$ ($i \ne j$). While $\bM_{n \times n}^{[1]}$ covers the smallest attention scope, $\bM_{n \times n}^{[K]}$ is a full social mask with all elements equal to 1. 
Then, we have the refined individual-individual interaction:
\begin{align}
    \bh_{\mathbf{m}, t}^{[k]} &= \text{GAT}(\bM^{[k]}\bh_{t}^{[k]} ).
\end{align}
This scope-changing social mask is useful for pedestrian prediction, since a pedestrian cares more about its neighbors within a closer distance. However, in team sports, a player shall always pay social attention to his teammates and opponents. To this end, we set the social mask in team sports always as a matrix of ones $\bM_{ij}^{[k]}= 1, \forall i,j \in [1,n]$, where every element is equal to one.

\subsection{Loss Function} 
The total loss of IHVRNN consists of two parts: the loss in generation and the loss in prediction, respectively. 
\begin{align}
    \mathcal{L}_{loss} = \mathcal{L}_{gen} + \mathcal{L}_{pred},
\end{align}
where $\mathcal{L}_{gen}$ is negative evidence lower bound \cite{alemi2018fixing} and $\mathcal{L}_{pred}$ is the $L_2$-norm between the prediction and its corresponding ground-truth. For details, see supplementary materials.

\section{Experiment}
\label{expriment}


In this section, we evaluate our method on two benchmark tasks, \emph{i.e.}, pedestrian and team sports trajectory prediction, which provides \textbf{collaborative} and \textbf{competitive} scenarios with multiple agents. 
The experiment shows our method has more advantages in both competitive and cooperative scenarios for multi-agent prediction. The supplementary materials demonstrates some learned “tactics” and the controllable experiment of “group consensus (strategy)”. A demo video is also available.

To make a \textbf{fair} comparison to both uni-modal and multi-modal methods, we need to align the ``evaluation strategy". Generally, the uni-modal or deterministic methods learn average outcomes and generate a single output, while the multi-modal methods can learn $N$ possible outcomes under $M$ modals ($N$$<$$M$). The design of selecting $N$ outcomes may differ from each multi-modal method \cite{fang2020tpnet,zhao2020tnt,mangalam2020not,DenseTNT_gu2021densetnt,lee2017desire}, further making the comparison complicated. \cite{chen2021personalized} proposed a new evaluation strategy called the probability cumulative minimum distance (PCMD) curve to enable fair comparison. 
We therefore use the "single output" setting ($PCMD(1/M)$) \cite{chen2021personalized}) to facilitate the evaluation of the different methods, so that we can better focus on exploring how IGC benefits trajectory prediction.

\subsection{Team Sports}

\vspace{1ex}
\noindent{\textbf{Datasets~and~Evaluation~Metrics.}}~~~ 
The proposed method is tested on the real professional soccer league dataset
that has been used in \cite{bialkowski2014win} and \cite{le2017coordinated}. 
The dataset consists of tracking data from 45 {\it Europa League} matches \cite{le2017coordinated} at 10 frames per second. 
Following the same setting in \cite{le2017coordinated}, {\color{black} every} model is trained with the observation length of 50 frames and the prediction length of 10 frames, whereas its performance is evaluated on the prediction length of 10 and 30. Following the evaluation configuration in previous work, we evaluate the performance of our methods and other baselines on mean square errors (MSE): average squared $L_2$ distance between the ground truth and the prediction.


\vspace{1ex}
\noindent{\textbf{Baselines.}}~~~
We select several popular benchmark methods in team sports, which focus on authentic trajectories generations. While multiple agents in team sports have a strong interaction, the methods for pedestrian prediction \cite{alahi2016social,gupta2018social,huang2019stgat} with a lot of success in capturing the interaction between multi-agents are also suitable to address the prediction problem of team sports. We adapt several baselines for team sports, including RNN \cite{hochreiter1997long}, Dynamic RNN (DRNN), Role assignment with LSTMs~\cite{le2017coordinated}, Social LSTM \cite{alahi2016social}, Social GAN \cite{gupta2018social}, GVRNN \cite{yeh2019diverse} and STGAT~\cite{huang2019stgat}. 
\begin{table}[h!] 
\caption{Comparison with baseline methods on the real Europa League dataset for $T_{pre}=10$ and $30$. Each column represents a method and each row represents the MSE loss with different prediction lengths. The smaller value is better.}

\label{table_soccer}
\centering
\tiny{
\begin{tabular}{c|llllllllll}
\hline
Pred len &LSTM & DRNN& S-LSTM&S-GAN &	STGAT  & R-Assign	&GVRNN &VRNN &HVRNN &IHVRNN  \\
\hline
10 &0.0120 &0.0144&	0.0168&	0.0100 &0.0075 &0.0041&	0.0096 & 0.0091 & 0.0128& \textbf{0.0038}\\
30&	0.0999&	0.0977&	0.1231&	0.1165&	0.0880&0.1039&	0.0898 & 0.0979& 0.0829 & \textbf{0.0499}\\
\hline

\end{tabular}
}
\end{table}

\vspace{1ex}
\noindent{\textbf{Experiment Results.}}~~~
Table \ref{table_soccer} shows the results of our model and these baseline methods on the Europa League dataset. 
It can be found that our IHVRNN can outperform all the baselines (LSTM, Stochastic LSTM, VRNN, Role-Assign, Social-LSTM, Social-GAN, and STGAT) for both $T_{pred} =10$ and $30$, in term of MSE loss. 
To further verify the effectiveness of interactive group consensus (IGC), we also conduct the ablation study in the team sports task. 
The last three models (VRNN, HVRNN, IHVRNN) in Table \ref{table_soccer} shows 1) the model incorporated group consensuses (IHVRNN and HVRNN) can achieve better performance, and 2) the model with IGC considering multi-level dynamics (IHVRNN) 
outperforms the model only consider intra-group dynamics (HVRNN). This demonstrates the effectiveness of IGC.

\begin{figure}[!ht]
  \centering
   \begin{subfigure}[b]{0.47\linewidth}
         \centering
         \includegraphics[width=\textwidth]{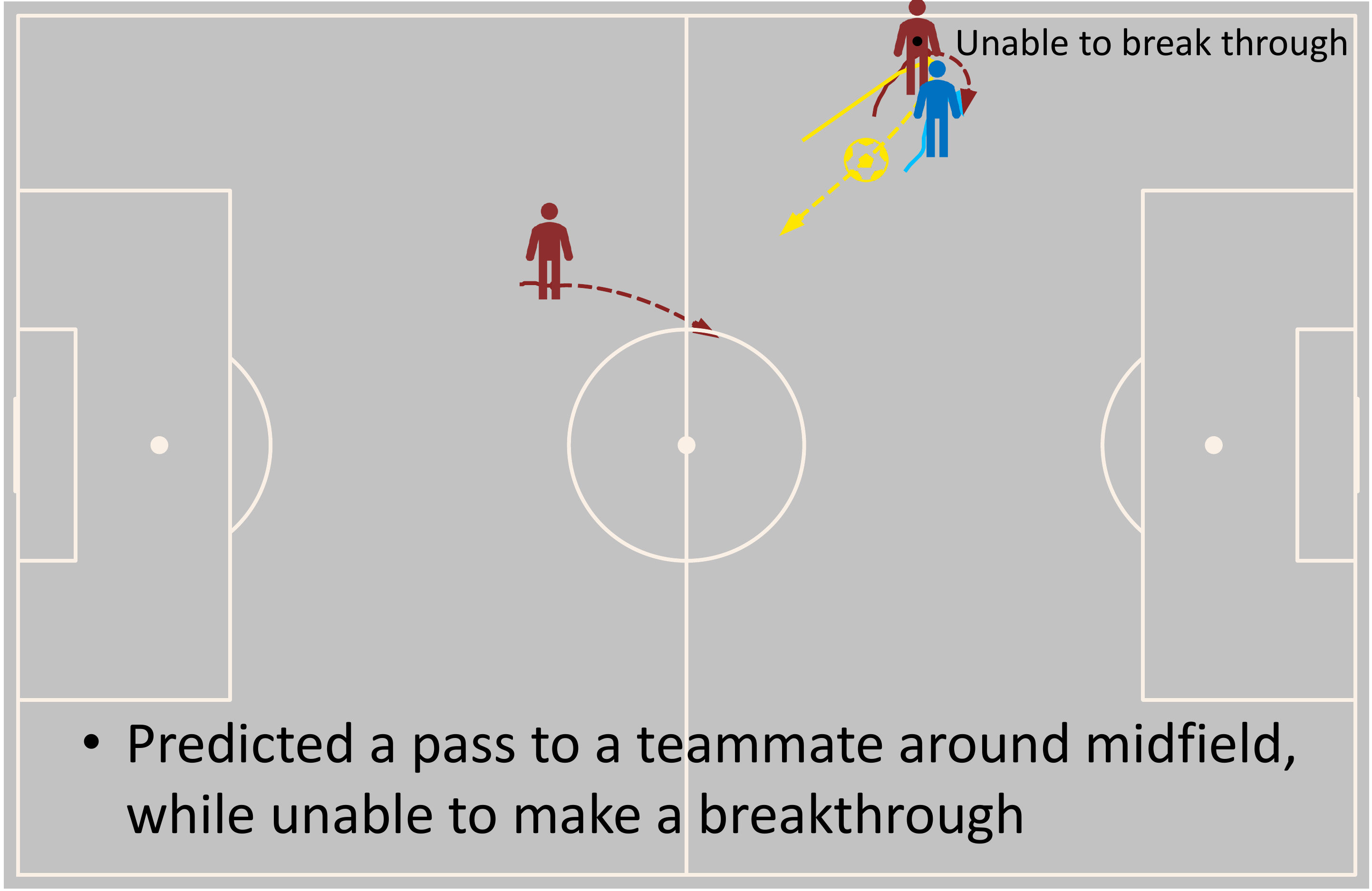}
         \caption{} 
         \label{fig:unbreakthrough-2}
    \end{subfigure}
    \begin{subfigure}[b]{0.47\linewidth}
         \centering
         \includegraphics[width=\textwidth]{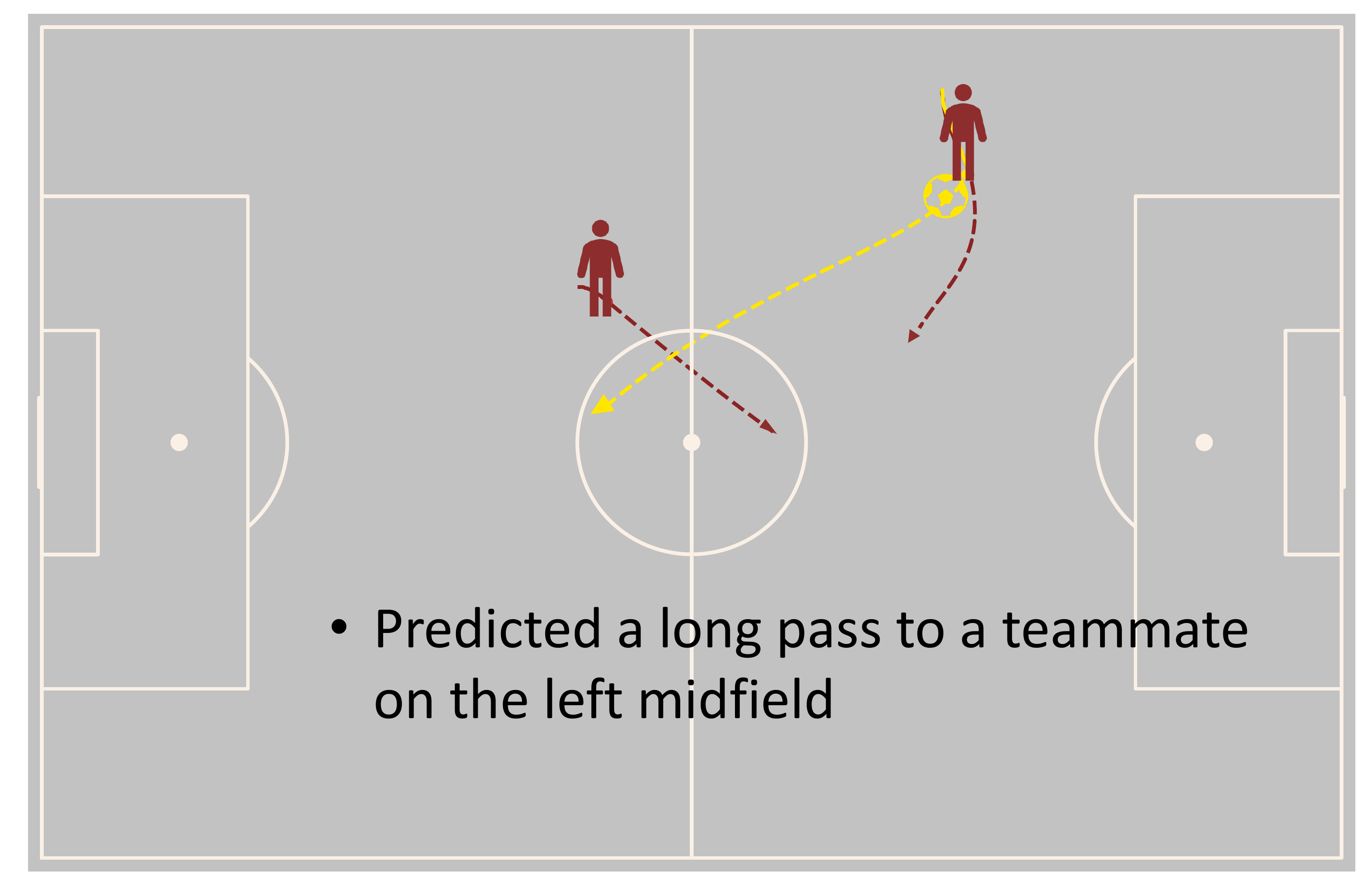}
         \caption{ }
         \label{fig:non-breakthrough}
     \end{subfigure}
  \vskip -0.2cm
  \caption{Example for learning attacking and passing tactics: (a) Predicting a pass to a teammate around midfield, when an offensive player is unable to make a breakthrough from the left-wing, and (b) Predicting a long pass to a teammate on the midfield.
}
  \label{fig:learning-tactics}
  \vspace{-0.2cm}
\end{figure} 

\vspace{1ex}
\noindent{\textbf{Qualitative study.}}~~~Our framework can showcase some tactics learning examples in soccer trajectory prediction, \emph{e.g.}, attacking and passing tactics (See Figure \ref{fig:learning-tactics}). More qualitative studies are given in supplementary materials. 
A demo video is also available, which showcases several examples for a long prediction time ($T_{pre}=50$).

\subsection{Pedestrian Trajectory Prediction}
\vspace{1ex}
\noindent{\textbf{Datasets.}}~~~
We adapt our framework to the pedestrian prediction on two popular benchmarks, ETH \cite{park2018sequence} and UCY \cite{lerner2007crowds}, which are widely used for human trajectory prediction. 
ETH and UCY contain five crowd scenes for pedestrian prediction: ETH, HOTEL, UNIV, ZARA1, and ZARA2. 
{\color{black} We follow the same configuration as in previous literature \cite{alahi2016social,gupta2018social,liang2019peeking,li2019way,sadeghian2019sophie,zhang2019sr,huang2019stgat,fang2020tpnet,sun2020recursive}, 
\emph{i.e.}, the model observes the past trajectories for 3.2 seconds (8 frames) and then predicts for 3.2 seconds (8 frames) and 4.8 seconds (12 frames), respectively.
}

\vspace{1ex}
\noindent{\textbf{Evaluation Metrics.}}~~~
Following the previous works,  
these two metrics are commonly used for evaluating the performance of pedestrian prediction:
{\textbf {Average Displacement Error}} (ADE) is the average $L_2$ distance between the ground truth and predicted trajectories,
and {\textbf{Final Displacement Error}} (FDE) is the $L_2$ distance between the ground truth destination and the predicted destination at the last prediction step. 

\begin{table*}[!t]
\caption{The comparison between our IHVRNN and interaction-based baselines under the aforementioned \textbf{fair} evaluation setting for $T_{pre}=12$: (ADE / FDE). The smaller value is better. 
} 
\vskip 0.1cm
\label{table:ped}
\centering
\begin{tabular}{l|c|c|c|c|c|c}
\hline
Methods      & ETH      & 	HOTEL    &	UNIV 	& 	ZARA1  &	ZARA2  & AVG \\
\hline
Social-LSTM  & 1.09/2.35& 0.79/1.76& 0.67/1.40 & 0.47/1.00& 0.56/1.17& 0.72/1.52\\
Social-GAN   & 0.98/1.98& 0.63/1.36& 0.64/\textbf{0.58} & 0.47/1.02& 0.39/0.87& 0.65/1.32\\
SoPhie       & \textbf{0.70}/\textbf{1.43}& 0.76/1.67& 0.53/1.24 & \textbf{0.30}/\textbf{0.63}& 0.38/0.78& 0.54/1.15\\ 
TPNet-1      & 1.00/2.01& 0.31/\textbf{0.46}& 0.55/0.94 & 0.46/0.75& 0.33/0.60& 0.53/\textbf{0.90}\\    
RSBG         & 0.79/1.47& 0.35/0.71& 0.68/1.39 & 0.42/0.89& 0.35/0.71& 0.52/1.03\\ 
GVRNN        & 0.92/2.05& 0.31/0.59& 0.52/1.16 & 0.41/0.89& 0.32/0.72& 0.50/1.08\\  
STGAT        & 1.03/2.20& 0.59/1.21& 0.68/1.49 & 0.53/1.17& 0.43/0.94& 0.65/1.40\\    
Social-STGCNN& 1.01/1.83& 0.74/1.42& 0.71/1.38 & 0.57/1.13& 0.51/0.97& 0.71/1.35\\    
Trajectron++ & 1.38/2.10& 0.63/0.88& 0.51/0.85 & 0.46/0.68& 0.30/\textbf{0.48}& 0.65/1.00\\ 
\hline
IHVRNN (Ours)& 0.83/1.81& \textbf{0.28}/0.53& \textbf{0.51}/1.12 & 0.39/0.85& \textbf{0.30}/0.66& \textbf{0.47}/1.00\\
\hline
\end{tabular}
\end{table*}

\begin{table}[!t]
\caption{The settings for the ablation study on the effectiveness of a different combination of components in IHVRNN: interactive group consensus (IGC), group consensus (GC), and refinement module (RM), and GAT. The GC shows the case without the group-group interaction, where a group does not speculate about the common strategy from the other groups.}
\label{table:ablation-setting}
\centering
\small{
\begin{tabular}{c|llll}
\hline
      & IGC & 	GC  & RM & GAT\\
\hline 
IHVRNN &   \checkmark  &   &  \checkmark &  \checkmark \\    
HVRNN &    &  \checkmark &  \checkmark &  \checkmark \\
VRNN &  & &   \checkmark &  \checkmark \\
IHVRNN (IGC+RM) &   \checkmark  &   &  \checkmark &\\
IHVRNN (IGC+GAT) &   \checkmark  &   &   & \checkmark\\
IHVRNN (IGC only) &   \checkmark  &   &   &\\
\hline
\end{tabular}
}
\end{table}

\begin{table}[!t]
\caption{
The ablation study on the effectiveness of the interactive group consensus (IGC), as shown in the first three rows of Table \ref{table:ablation-setting}. Both IHVRNN and HVRNN contain group consensuses (GC), whereas IHVRNN preserves the reasoning on the other groups' consensuses (contains  IGC) and HVRNN only contains intra-group consensus without conjecture about the others' group consensuses. VRNN shows the case without GC. (Each entry: $T_{pre}$=8/$T_{pre}$=12.)}
\setlength{\tabcolsep}{3.6pt}
\label{table:ped-ablation-IGC}
\centering
\small{
\begin{tabular}{c|llll}
\hline
Metric & Dataset & 	IHVRNN  & HVRNN& VRNN\\
\hline
\hline
 & ETH &  \textbf{0.499 / 0.830} & 0.538 / 0.897 & 0.541 / 0.900\\
 & HOTEL& \textbf{0.195 / 0.283} & 0.199 / 0.290 & 0.244 / 0.385\\
ADE & UNIV& 0.300 / 0.513 &	0.302 / 0.514 & \textbf{0.295 / 0.502}\\
   & ZARA1  & 	\textbf{0.226 / 0.388} & 0.234 / 0.399 & 0.233 / 0.397\\
&ZARA2&	\textbf{0.178 / 0.303} &	0.180 / 0.306 & 0.187 / 0.324\\    
\hline
AVG & &\textbf{0.280 / 0.463}& 0.290 / 0.481 & 0.300 / 0.502\\
\hline
& ETH 	&\textbf{0.977 / 1.814} &1.063 / 1.960 &1.072 / 1.963\\
&HOTEL &\textbf{0.337 / 0.532}	& 0.346 / 0.550 & 0.457 / 0.790\\
FDE&UNIV &	0.637 / 1.122	 &0.640 / 1.125 & \textbf{0.624 / 1.096}\\
&ZARA1 	&\textbf{0.471 / 0.855}&	0.492 / 0.884& 0.487 / 0.876\\
&ZARA2	&\textbf{0.374 / 0.659} 	&0.380 / 0.665 & 0.397 / 0.718\\
\hline
AVG& &\textbf{0.560 / 0.996}& 0.584 / 1.037& 0.607 / 1.089\\
\hline
\end{tabular}
}

\end{table}
\begin{table}[!t]
\caption{
The ablation study on the effectiveness of the effectiveness of the refinement module (RM) and the GAT (as shown in the last three rows in Table \ref{table:ablation-setting}). 
In particular, IGC+RM represents our method with an RM but without GAT, whereas IGC+GAT shows the version without an RM. IGC only means the model using neither RM nor GAT. (Each entry: $T_{pre}$=8/$T_{pre}$=12.) 
}
\vskip 0.1cm
\setlength{\tabcolsep}{3.6pt}
\label{table:ped-ablation-RM-GAT}
\centering
\small{
\begin{tabular}{c|llll}
\hline
Metric & Dataset &  IGC+RM  & IGC+GAT & IGC only\\
\hline
\hline
 & ETH &    0.549 / 0.926 & {\textbf{0.501 / 0.842}}& 0.514 / 0.826\\
 & HOTEL&  0.208 / 0.302 & {\textbf{0.204 / 0.297}} & 0.241 / 0.383\\
ADE & UNIV&  {\textbf{0.300}} / 0.516 &	0.302 / {\textbf{0.514}} & 0.352 / 0.596\\
   & ZARA1   & 0.241 / 0.413 & {\textbf{0.235 / 0.402}} & 0.261 / 0.434\\
&ZARA2&	0.188 / 0.328 &	{\textbf{0.181 / 0.308}} & 0.238 / 0.405\\    
\hline
AVG & & 0.297 / 0.497& {\textbf{0.285 / 0.473}} & 0.321 / 0.529\\
\hline
& ETH 	&1.097 / 2.037 & 0.993 / 1.857 &{\textbf{0.978 / 1.702}}\\
&HOTEL  &0.366 / 0.562	& {\textbf{0.353 / 0.558}} & 0.462 / 0.792\\
FDE&UNIV&	0.640 / 1.138	 &{\textbf{0.641 / 1.123}} & 0.727 / 1.299\\
&ZARA1 	&0.511 / 0.914&	{\textbf{0.493 / 0.887}}& 0.527 / 0.936\\
&ZARA2	&0.405 / 0.730 	&{\textbf{0.380 / 0.673}} & 0.493 / 0.882\\
\hline
AVG& & 0.604 / 1.076& {\textbf{0.572 / 1.020}}& 0.637 / 1.122\\
\hline
\end{tabular}
}
\end{table}

\vspace{1ex}
\noindent{\textbf{Baselines.}}~~~
We compare our IHVRNN with several competitive methods which model the interaction among pedestrians: Social LSTM \cite{alahi2016social}, Social GAN \cite{gupta2018social}, SoPhie \cite{sadeghian2019sophie}, STGAT \cite{huang2019stgat}, RSBG \cite{sun2020recursive}, TPNet-1 \cite{fang2020tpnet}, Social-STGCNN \cite{mohamed2020social}, and Trajectron++ \cite{salzmann2020trajectron++}.
{We also re-implement GVRNN \cite{yeh2019diverse}, a state-of-the-art method for team sports prediction, on pedestrian prediction.} To better evaluate the effectiveness of different interaction modeling, goal-based methods \cite{mangalam2020not,mangalam2021goals} are not included in this comparison.

\vspace{1ex}
\noindent{\textbf{Experiment~Results.}}~~~
The experiment results of pedestrian trajectory prediction are summarized in Table \ref{table:ped}, where each entry shows the values of ADE and FDE. On average evaluation, it can be found that our IHVRNN achieves better results compared to all these interaction-based baselines according to ADE. Comparing to Trajectron++ (the SOTA stochastic method), our method always has the smallest ADE in all these five datasets. The rankings of our method in these datasets are: $3^{rd}$, $1^{st}$, $1^{st}$, $2^{st}$, $1^{st}$ on ADE, or $3^{rd}$, $2^{nd}$, $4^{th}$, $4^{th}$, $3^{rd}$ on FDE. 
Generally, the performance of these interaction-based baselines only stands out only in 1 or 2 datasets. One possible reason is that: the dominant element of pedestrians' interaction is changing among different datasets and is complicated to capture consistently by existing methods. Comparing to these baselines, our method can show consistent result in most of the datasets according to ADE.


\vspace{1ex}
\noindent{\textbf{Ablation~Study.}}~~~
To further validate the effectiveness of different components in our method, we design several variants to perform the ablation. The detailed settings of these variants are shown in Table~\ref{table:ablation-setting}, with different combinations of group consensus (GC), RM, and GAT. 

\vspace{1ex}
\noindent{\textbf{1) Effect of interactive group consensus (IGC).}} We conduct ablation on several variants, including IHVRNN, HVRNN, and VRNN, to investigate the effectiveness of IGC. {Results of the ablation study in Table \ref{table:ped-ablation-IGC} show IHVRNN performs best among the three methods. Compared to VRNN (without exploring GC), HVRNN (with GC and its intra-group dynamics) has a better performance, which demonstrates the benefit of GC. Furthermore, IHVRNN with IGC is better than HVRNN with GC, further demonstrates the effectiveness of IGC.}


\vspace{1ex}
\noindent{\textbf{2) Effect of GAT and RM.}}
The last three columns, in Table \ref{table:ped-ablation-RM-GAT}, show the effectiveness of GAT and the refinement module in our framework. {In Table \ref{table:ped-ablation-RM-GAT}, we design several variants, including RM+IGC, GAT+IGC, and IGC, to demonstrate the effectiveness of using GAT or RM. The results indicate that both RM \& GAT make contributions in the final performance, because using IGC+RM or IGC+GAT performs better than using IGC only.}




\vspace{1ex}
\noindent{\textbf{Discussion.}} Why IGC performs better in team sports rather than pedestrians datasets? Firstly, our method tries to model how group consensuses evolve and how they influence their group members in multi-agent trajectory prediction. Additionally, team sports give high collaborative and competitive scenarios and naturally have two teams and each team has a specific strategy. Thus, the interactive group consensus better models this phenomenon in team sports.

\section{Conclusions and future work}


In this paper, we propose a novel variational framework for human trajectory prediction by introducing interactive group consensus (IGC) as a new concept. The key idea of our method is to define and model IGC in its corresponding interactive hierarchical latent space, with the help of which our method can capture various interactions on multiple hierarchies. 
The better performance on different types of datasets indicates the good generalizability and effectiveness of IGC. 
In particular, our method is more suitable for highly collaborative and competitive multi-agent scenarios. In future work, we will evaluate our method on larger-scale sports datasets once available.  

\clearpage
%
%
\bibliographystyle{splncs04}
\bibliography{eccv_11.bib}
\end{document}